\documentclass{article}
\usepackage{spconf,amsmath,graphicx}
\usepackage{balance}
\usepackage{multirow}

\title{Level up the deepfake detection: a method to effectively discriminate images generated by GAN architectures and Diffusion Models}
\name{Luca Guarnera$^{\star}$, Oliver Giudice$^{\dagger}$, Sebastiano Battiato$^{\star}$}
\address{$^{\star}$ Department of Mathematics and Computer Science, University of Catania, Italy\\
$^{\dagger}$ Applied Research Team, IT dept., Banca d'Italia, Rome, Italy\\
\emph{\{luca.guarnera, sebastiano.battiato\}@unict.it, oliver.giudice@bancaditalia.it}}

\begin{document}
%
\maketitle
\begin{abstract}
The image deepfake detection task has been greatly addressed by the scientific community to discriminate real images from those generated by Artificial Intelligence (AI) models: a binary classification task. In this work, the deepfake detection and recognition task was investigated by collecting a dedicated dataset of pristine images and fake ones generated by $9$ different Generative Adversarial Network (GAN) architectures and by $4$ additional Diffusion Models (DM). 

A hierarchical multi-level approach was then introduced to solve three different deepfake detection and recognition tasks: (i) Real Vs AI generated;  (ii) GANs Vs DMs; (iii) AI specific architecture recognition. Experimental results demonstrated, in each case, more than 97\% classification accuracy, outperforming state-of-the-art methods.

\end{abstract}
\section{Introduction}
\label{sec:intro}

\begin{figure*}[t!]
    \centering     \includegraphics[width=\linewidth]{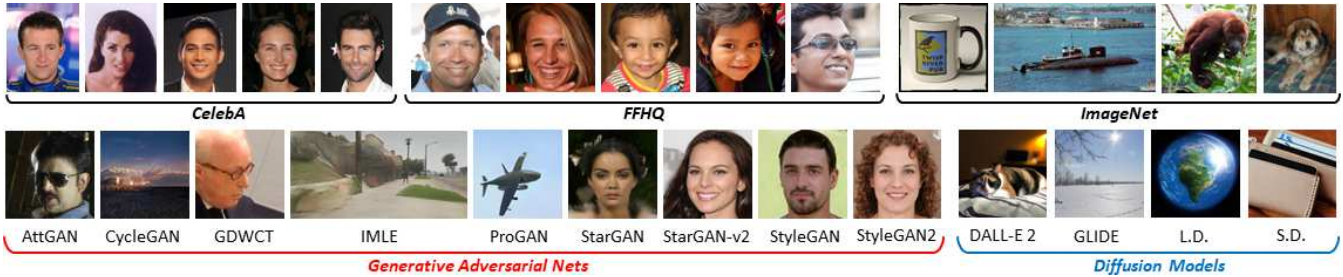}
    \vspace{-0.5cm}
    \caption{Examples of images collected from different datasets and images generated by different GANs and DMs.}
    \label{fig:db}
\end{figure*}

The term deepfake refers to all those multimedia contents generated an AI model. The most common deepfake creation solutions are those based on GANs~\cite{goodfellow2014generative} which are effectively able to create from scratch or manipulate a multimedia data. In a nutshell, GANs are composed by two neural networks: the \textit{Generator (G)} and the \textit{Discriminator (D)}. \textit{G} creates new data samples that resemble the training data, while \textit{D} evaluates whether a sample is real (belonging to the training set) or fake (generated by the \textit{G}). A GAN must be trained until \textit{D} is no longer able to detect samples generated by \textit{G}, in other words, when \textit{D} starts to be fooled by \textit{G}.
Several surveys on methods dealing with GAN-based approaches for the creation and detection of deepfakes, have been proposed in~\cite{masood2022deepfakes,nowroozi2021survey}.

Recently, DMs~\cite{sohl2015deep,ho2020denoising} are arousing interest thanks to their photo-realism and also to a wide choice in output control given to the user. In contrast to GANs, DMs are a class of probabilistic generative models that aims to model complex data distributions by iteratively adding noise to a random noise vector input for the generation of new realistic samples and, using them as basis, proceed to reconstruct the original data.
Stable Diffusion~\cite{rombach2022high} and DALL-E 2~\cite{ramesh2022hierarchical} are the most famous state of the art DMs, based on the text-to-image translation operation. 
As demonstrated in~\cite{dhariwal2021diffusion}, DMs are able to produce even better realistic images than GANs, since GANs generate high-quality samples but are demonstrated to fail in covering the entire training data distribution. 

To effectively counteract the illicit use of synthetic data generated by GANs and DMs, new deepfake detection and recognition algorithms are needed.
As far as image deepfake detection methods in state of the art are concerned, they mostly focus on binary detection (Real Vs. AI generated \cite{zhang2019detecting,wang2019fakespotter}) . Interesting methods in state of the art already demonstrated to effectively discriminate between different GAN architectures \cite{Giudice_JI_2021,wang2020cnn,guarnera2020fighting}. Methods to detect DMs and recognize them have been proposed just recently \cite{sha2022fake,corvi2022detection}. 

In order to level up the deepfake detection and recognition task, the objective of this paper and the main contribution is to classify an image among $14$ different classes: $9$ GAN architectures, $4$ DMs engines and $3$ pristine datasets (labeled as belonging to the same ``real" class). At first, a dedicated dataset of images was collected. 
Then, a novel multi-level hierarchical approach exploiting ResNET models was developed and trained. The proposed approach consists of 3 levels of classification: (Level 1) Real Vs AI-generated images; (Level 2) GANs Vs DMs; (Level 3) recognition of specific AI (GAN/DM) architectures among those represented in the collected dataset. 
Experimental results demonstrated the effectiveness of the proposed solution, achieving more than $97\%$ accuracy on average for each task, exceeding the state of the art. Moreover, the hierarchical approach can be used to analyze multimedia data in depth to reconstruct its history (forensic ballistics) \cite{guarnera2022deepfake}, a task poorly addressed by the scientific community on synthetic data.

This paper is organized as follows: Section \ref{sec:dataset} and Section \ref{sec:method} describe the dataset and the proposed approach built upon it respectively. Experimental results and comparison are presented in Section \ref{sec:exp}. Finally, Section \ref{sec:conclusion} concludes the paper.

\section{Dataset details}
\label{sec:dataset}
The dataset employed in this study is a dedicated collection of images: real/pristine images collected from CelebA \cite{liu2015deep}, FFHQ\footnote{https://github.com/NVlabs/ffhq-dataset}, and ImageNet \cite{russakovsky2015imagenet} datasets and synthetic data generated by $9$ different GAN engines (AttGAN \cite{he2019attgan}, CycleGAN \cite{zhu2017unpaired}, GDWCT \cite{cho2019image}, IMLE \cite{li2019diverse}, ProGAN \cite{karras2017progressive}, StarGAN~\cite{choi2018stargan}, StarGAN-v2~\cite{choi2020stargan}, StyleGAN~\cite{karras2019style}, StyleGAN2~\cite{karras2020analyzing}) and $4$ text-to-image DM architectures (DALL-E 2~\cite{ramesh2022hierarchical}, GLIDE~\cite{nichol2021glide}, Latent Diffusion~\cite{rombach2022high}\footnote{a.k.a. Stable Diffusion: https://github.com/CompVis/stable-diffusion}. For each considered GAN, $2,500$ images (a total of $22,500$) were generated while for the DMs, $5,000$ images were created for each architecture employing more than $800$ random sentences, for a total of $20,000$ images. Overall, the total number of synthetic data consists of $42,500$ images. Finally, for each real dataset (CelebA, FFHQ and ImageNet) $13,500$ images were considered, for a total of $40,500$. 

Table \ref{tab:datasetdetails} summarizes the numbers of the obtained dataset with respect to each level and to division of training, validation and test sets employed for the experiments. 
Figure \ref{fig:db} shows several examples of the obtained dataset.

\begin{table}[t!]
\small
\centering
\begin{tabular}{ccc|c|c|c}
\hline
\multicolumn{3}{c|}{\textbf{\begin{tabular}[c]{@{}c@{}}Classification\\ Task\end{tabular}}}                                           & \textbf{\begin{tabular}[c]{@{}c@{}}Train \\ 50\%\end{tabular}} & \textbf{\begin{tabular}[c]{@{}c@{}}Val\\ 20\%\end{tabular}} & \textbf{\begin{tabular}[c]{@{}c@{}}Test \\ 30\%\end{tabular}} \\ \hline
\multicolumn{2}{c}{\multirow{2}{*}{\textbf{14-classes}}}       & \textbf{Total Images}                                                  & 28,000                                                        & 4.200                                                      & 7.000                                                        \\ \cline{3-6} 
\multicolumn{2}{c}{}                                         & \textbf{\begin{tabular}[c]{@{}c@{}}\#Img $\forall$ class\end{tabular}} & 2,000                                                         & 300                                                        & 500                                                          \\ \hline
\multicolumn{2}{c}{\multirow{2}{*}{\textbf{13-classes}}}       & \textbf{Total Images}                                                  & 26,000                                                        & 3,900                                                      & 6,500                                                        \\ \cline{3-6} 
\multicolumn{2}{c}{}                                         & \textbf{\begin{tabular}[c]{@{}c@{}}\#Img $\forall$ class\end{tabular}} & 2,000                                                         & 300                                                        & 500                                                          \\ \hline
\multicolumn{2}{c}{\multirow{2}{*}{\textbf{L1}}}             & \textbf{Total Images}                                                  & 46,480                                                        & 11,620                                                     & 24,900                                                       \\ \cline{3-6} 
\multicolumn{2}{c}{}                                         & \textbf{\begin{tabular}[c]{@{}c@{}}\#Img $\forall$ class\end{tabular}} & 23,240                                                        & 5,810                                                      & 12,450                                                       \\ \hline
\multicolumn{2}{c}{\multirow{2}{*}{\textbf{L2}}}             & \textbf{Total Images}                                                  & 23,800                                                        & 5,950                                                      & 12,750                                                       \\ \cline{3-6} 
\multicolumn{2}{c}{}                                         & \textbf{\begin{tabular}[c]{@{}c@{}}\#Img $\forall$ class\end{tabular}} & 11,900                                                        & 2,975                                                      & 6,375                                                        \\ \hline
\multirow{4}{*}{\textbf{L3}} & \multirow{2}{*}{\textbf{GANs}} & \textbf{Total Images}                                                  & 12.600                                                        & 3,150                                                      & 6,750                                                        \\ \cline{3-6} 
                             &                                & \textbf{\begin{tabular}[c]{@{}c@{}}\#Img $\forall$ class\end{tabular}} & 1,400                                                         & 350                                                        & 750                                                          \\ \cline{2-6} 
                             & \multirow{2}{*}{\textbf{DMs}}  & \textbf{Total Images}                                                  & 11,200                                                        & 2,800                                                      & 6,000                                                        \\ \cline{3-6} 
                             &                                & \textbf{\begin{tabular}[c]{@{}c@{}}\#Img $\forall$ class\end{tabular}} & 2,800                                                         & 700                                                        & 1,500                                                        \\ \hline
\end{tabular}
\caption{Overview of the images employed for training, validation and test sets (last three columns with indication of \% of samples). The first column denotes the classification task (e.g., $14-classes$ is the flat classification task with 14 classes; $L1$ refers to the Level 1 of hierarchy. The $Total  Images$ rows indicate the total number of images employed for training, validation, and testing phases. The $\#Img \forall class$ represents the number of samples considered for each class.}
\label{tab:datasetdetails}
\vspace{-0.45cm}
\end{table}

\section{Multi-level Deepfake Detection and Recognition}
\label{sec:method}
\begin{figure*}[t!]
    \centering 
    \includegraphics[width=\linewidth]{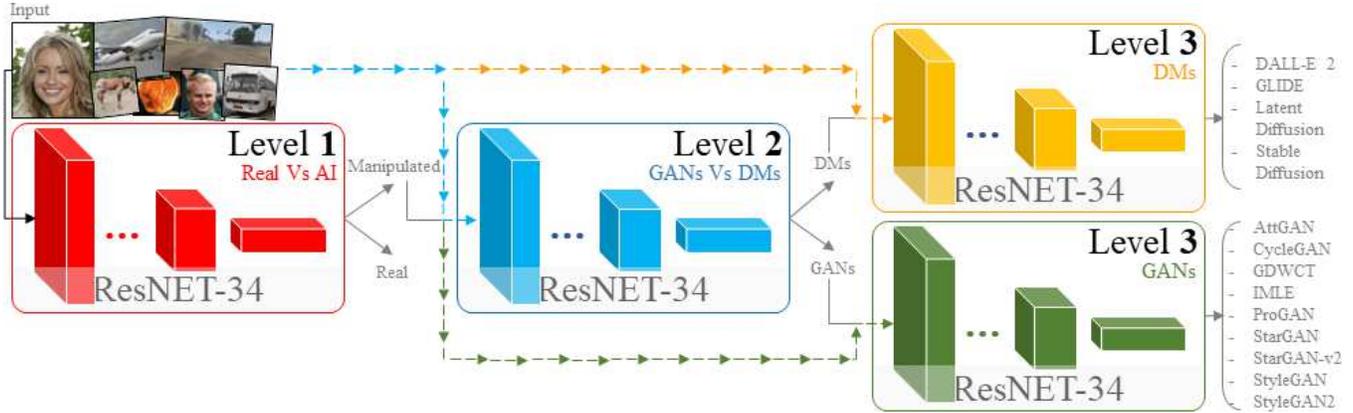}
    \vspace{-0.5cm}
    \caption{Execution flow of the proposed hierarchical approach. Level 1 classifies images as real or AI-generated. Level 2 defines whether the input images were created by GAN or DM technologies. Level 3, composed of two sub-modules, solves the AI architecture recognition task. The dashed arrows represent an optional flow (e.g., in the case the input image is real, it will not be analyzed by the next levels).
    }
    \label{fig:proposedpipeline}
\end{figure*}

The dataset collected and described in the previous section was preliminarly investigated as a ``flat" classification task. This was carried out by employing a single ResNET-34 encoder with $14$ classes as output layer. In addition, a further test was carried out by removing every image belonging to the real class ($13$ classes as output layer). The trained ResNET-34 model showed that in this last case it is able to achieve greater accuracy score. This gave the idea that a hierarchical approach could lead to even better results also giving a bit of explainability on the analyzed image.

The proposed multi-level deepfake detection and recognition approach consists of 3 levels.  Level 1 has the objective to detect real data from those created AI architectures (so all synthetic data were labeled as belonging to the same class). Thus this level is implemented as a binary ResNET-34 classifier. Given that an image was previously classified as generated by an AI, Level 2 furtherly analyzes images to discriminate between those generated by a GAN from those generated by a DM. Thus this level is implemented as another binary ResNET-34 classifier. 
Finally, given that an image was previously classified as generated by a GAN or by a DM, the last level solves the task of recognizing the specific architecture between those considered in the dataset described in Section \ref{sec:dataset}. To do this, Level 3 is divided into two sub-modules: ``Level 3-GANs" to discriminate the specific GAN, implemented as a 9-classes ResNET-34 classifier; and ``Level 3-DMs" implemented as 4-classes ResNET-34 Diffusion Model classifier. Figure \ref{fig:proposedpipeline} summarizes the overall approach.

\section{Experimental Results and Comparison}
\label{sec:exp}


Experiments with ResNET architecture were performed considering the following parameters for training: $batch size = 30$, $learning rate = 0.00001$, optimizer Stochastic Gradient Descent (SGD) with $momentum=0.9$ and Cross Entropy Loss. All images were resized to a resolution of $256x256$. As regards 14-classes, 13-classes and Level 1 models were trained for $150$ epochs. The remaining three models (one for Level 2 and two for Level 3 classification) were trained for $100$ epochs. This difference is due to the different amount of images at corresponding level (see Table \ref{tab:datasetdetails}).

Four different instances of the ResNET encoder were employed for the proposed multi-level deepfake detection and recognition task. Each ResNET model was properly trained with a corresponding sub-dataset composed as shown in Table \ref{tab:datasetdetails}. In particular, the Pytorch implementation of ResNET and each training started from the weights pre-trained on ImageNet\footnote{https://pytorch.org/vision/stable/models.html}. 
For each model, a fully connected layer with an output size equal to the number of classes of the corresponding classification level followed by a SoftMax was added to the last layer of ResNET encoders.

As far as model selection is concerned, ResNET architecture was chosen among other as already demonstrated to be effective in higher level deepfake recognition tasks \cite{guarnera2022exploitation}. Given the presented solution, $4$ models have to be run simultaneously on a single GPU. This limited the dimension of the useable models. Thus only ResNET-18 and ResNET-34 were took into account. The ResNET-34 architecture was selected as final solution for achieving the best results (see Table \ref{tab:compResnet}). 

At first, the ``flat" 14-classes classification task obtained an overall accuracy of only $94,85\%$ in the test-set. In order to slightly improve this result, first class (pristine images) was removed and the model retrained.  With the remaining 13 classes (only synthetic images generated by all considered GAN and DM engines) a bit better overall accuracy of $95,4\%$ was obtained. 
Taking advantage of this results, the hierarchical approach described in Section \ref{sec:method} was developed. 
The best results obtained are the following:
\begin{itemize}
\vspace{-0.2cm}
    \item  Level 1: classification accuracy of $97,63\%$;
    \vspace{-0.2cm}
    \item Level 2: classification accuracy of $98,01\%$;
    \vspace{-0.2cm}
    \item Level 3: an accuracy of $97,77\%$ was obtained for the GAN recognition task and an accuracy of $98,02\%$ for the DM one.
\end{itemize}
\begin{figure*}[t!]
    \centering 
    \includegraphics[width=\linewidth]{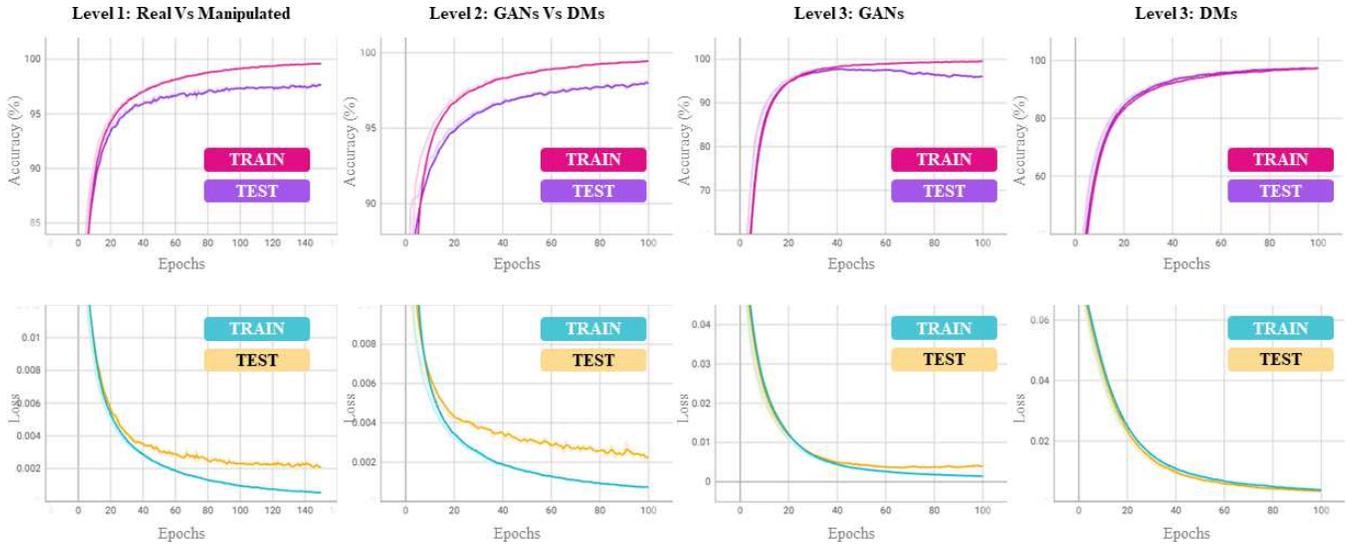}
    \vspace{-0.5cm}
    \caption{Trend of accuracy and error values obtained in training and test phases for each epoch of the three levels of the proposed hierarchical approach. Each column represents a classification level.} 
    \label{fig:acc_gerarchico}
\end{figure*}
Figure~\ref{fig:acc_gerarchico} shows the accuracy and error trends obtained in the training and test phases for each epoch and for each model of the hierarchical approach. Table \ref{tab:compResnet} shows that ResNET-34 achieves the best results compared with ResNET-18.

\begin{table}[t!]
\small
\centering
\begin{tabular}{cc|c|cc}
\cline{2-5}
\multicolumn{1}{l}{}           & \textbf{\textbf{Level 1}} & \textbf{Level 2}     & \multicolumn{2}{c}{\textbf{Level 3}}                \\ \cline{2-5} 
                                & \textbf{Real Vs AI}       & \textbf{GANs Vs DMs} & \multicolumn{1}{c|}{\textbf{GANs}}  & \textbf{\textbf{DMs}}   \\ \hline
\multicolumn{1}{c|}{ResNET-18} &                      95,24     & 97,68       & \multicolumn{1}{c|}{95,54} & 97.29          \\ \hline
\multicolumn{1}{c|}{ResNET-34}       & \textbf{97,63}            & \textbf{98,01}       & \multicolumn{1}{c|}{\textbf{97,77}} & \textbf{98,02} \\ \hline
\end{tabular}
\vspace{-0.3cm}
\caption{Comparison of classification accuracy value (\%) obtained between ResNET-18 and ResNET-34 architectures with respect to the proposed hierarchical approach.}
\label{tab:compResnet}
\vspace{-0.3cm}
\end{table}

The final solution was compared with various state-of-the-art works (\cite{zhang2019detecting,wang2019fakespotter,guarnera2020fighting,Giudice_JI_2021,wang2020cnn,sha2022fake}) demonstrating to achieve best results in each task. In Table \ref{tab:comp}, the Level 3 DMs column does not show some values given that corresponding approaches does not cover the corresponding task. Therefore, most of the methods do not seem to be able to effectively distinguish  between different DMs. This is because DMs leave different  traces on synthetic images than those generated by GAN engines. This claim is further empirically demonstrated by the results reported in Table \ref{tab:comp} on the ``Real Vs AI" column, where, due to the presence of DMs-generated images, the classification accuracy values of the various methods are sensibly lower than the $97,63\%$ obtained by the proposed approach. For this reason, a column regarding Level 2 task was not added, as the classification results of the various methods turn out to be significantly lower (almost random classifiers) than the classification accuracy of $98,01\%$ obtained by the proposed approach. Finally, all state of the art methods, including the proposed one are able to generalize well in terms of distinguishing between GAN architectures that created the synthetic data (GANs column).

\begin{table}[t!]
\small
\centering
\begin{tabular}{cc|cc}
\cline{2-4}
\multicolumn{1}{l}{}                      & \textbf{Level 1}             & \multicolumn{2}{c}{\textbf{Level 3}}                \\ \cline{2-4} 
                                           & \textbf{Real Vs AI} & \multicolumn{1}{c|}{\textbf{GANs}}  & \textbf{DMs}   \\ \hline
\multicolumn{1}{c|}{AutoGAN \cite{zhang2019detecting}}     & 68.5                         & \multicolumn{1}{c|}{80,3}           & ---            \\ \hline
\multicolumn{1}{c|}{Fakespotter~\cite{wang2019fakespotter}} & 74,22                        & \multicolumn{1}{c|}{95,32}          & ---            \\ \hline
\multicolumn{1}{c|}{EM~\cite{guarnera2020fighting}} & 86.57                        & \multicolumn{1}{c|}{95,02}          & ---            \\ \hline
\multicolumn{1}{c|}{DCT\cite{Giudice_JI_2021}} & 87,20                        & \multicolumn{1}{c|}{95,89}          & ---            \\ \hline
\multicolumn{1}{c|}{Wang et al. \cite{wang2020cnn}}   & 78.54                        & \multicolumn{1}{c|}{97.32}          & ---            \\ \hline
 \multicolumn{1}{c|}{DE-FAKE \cite{sha2022fake}}         & 90,52                        & \multicolumn{1}{c|}{--}             & 93,45          \\ \hline
\multicolumn{1}{c|}{Our}         & \textbf{97,63}               & \multicolumn{1}{c|}{\textbf{97,77}} & \textbf{98,02} \\ \hline
\end{tabular}
\vspace{-0.2cm}
\caption{Comparison with state of the art approaches. Classification accuracy value is reported (\%). --- are reported where the method can not handle the corresponding task. }
\label{tab:comp}
\vspace{-0.5cm}
\end{table}

\section{Conclusions}
\label{sec:conclusion}
In this paper, a deepfake detection and recognition solution is presented. The proposed solution is able to recognize if an image was generated among $9$ different GAN engines and $4$ DM models. 
However, the proposed solution demonstrated to outperform state of the art in all tasks. 

Future experiments could evaluate the robustness of the proposed method in real-world contexts (JPEG compression, scaling, etc.). 
Also, the possibility to identify some analytical traces \cite{corvi2022detection,Giudice_JI_2021,guarnera2020fighting} will be investigated just to exploit unusual statistics embedded on images generated by DMs.


\balance
\bibliographystyle{IEEEbib}
\bibliography{main}

\end{document}